\documentclass[letterpaper, 10 pt, conference]{ieeeconf}  % Comment this line out if you need a4paper

\IEEEoverridecommandlockouts  % This command is only needed if 
\usepackage[linesnumbered,boxed]{algorithm2e}
\usepackage{multirow}
\usepackage{booktabs}
\usepackage{subfigure}
\usepackage{color}
\usepackage{xcolor}
\usepackage[colorlinks,linkcolor=blue,citecolor=black]{hyperref}
\usepackage{graphicx}

\usepackage{etoolbox}
\makeatletter
\title{\LARGE \bf
LAFFNet: A Lightweight Adaptive Feature Fusion Network for Underwater Image Enhancement
}

\author{Hao-Hsiang Yang$^{1}$, Kuan-Chih Huang$^{1}$ and Wei-Ting Chen$^{1,2}$
\thanks{$^{1}$ASUS Intelligent Cloud Services, Asustek Computer Inc. Taipei, 112019, Taiwan {\tt\small (danny1$_-$yang, kuanchih$_-$huang, jimmy10$_-$chen)@asus.com}}%
\thanks{$^{2}$Graduate Institute of Electronics Engineering, National Taiwan University, Taipei
106, Taiwan}%
}

\begin{document}

\maketitle
\thispagestyle{empty}
\pagestyle{empty}

%%%%%%%%%%%%%%%%%%%%%%%%%%%%%%%%%%%%%%%%%%%%%%%%%%%%%%%%%%%%%%%%%%%%%%%%%%%%%%%%
\begin{abstract}

Underwater image enhancement is an important low-level computer vision task for autonomous underwater vehicles and remotely operated vehicles to explore and understand the underwater environments. Recently, deep convolutional neural networks (CNNs) have been successfully used in many computer vision problems, and so does underwater image enhancement. There are many deep-learning-based methods with impressive performance for underwater image enhancement, but their memory and model parameter costs are hindrances in practical application. To address this issue, we propose a lightweight adaptive feature fusion network (LAFFNet). The model is the encoder-decoder model with multiple adaptive feature fusion (AAF) modules. AAF subsumes multiple branches with different kernel sizes to generate multi-scale feature maps. Furthermore, channel attention is used to merge these feature maps adaptively. Our method reduces the number of parameters from 2.5M to 0.15M (around 94\% reduction) but outperforms state-of-the-art algorithms by extensive experiments. Furthermore, we demonstrate our LAFFNet effectively improves high-level vision tasks like salience object detection and single image depth estimation.
\end{abstract}

%%%%%%%%%%%%%%%%%%%%%%%%%%%%%%%%%%%%%%%%%%%%%%%%%%%%%%%%%%%%%%%%%%%%%%%%%%%%%%%%
\section{INTRODUCTION}

Underwater image enhancement aims to restore clear images from underwater images and is a challenging task because underwater images usually suffer from severe quality degradation due to light absorption and scattering in the water medium. Additionally, visually-guided robots and autonomous underwater vehicles rely on this enhancement technique to observe regions of interest for some high-level computer vision tasks like underwater docking \cite{kimball2018artemis}, an inspection of submarine cables and wreckage \cite{bingham2010robotic}, salience objection detection \cite{qin2019basnet}, and other operational decisions effectively, as shown in Fig. 1.
According to \cite{akkaynak2018revised}, \cite{akkaynak2019sea}, the physical model of an underwater image can be described as:
\begin{equation}
%\vspace{+0.5cm}
I_c = J_ce^{-\beta_c^{D}(v_{D})_z}+B_c^{\infty}(1-e^{-\beta_c^{B}(v_{B})_z})
%I(x)=F(J(x))
\label{eq:1}
%\vspace{-0.5cm}
\end{equation}
where $c$ represents each of the RGB color channels, $I_c$ is the captured image in underwater mediums, $J_c$ is the clear image that needs to be recovered, $z$ is the imaging range, $B_c^\infty$ the wideband veiling light; $\beta_c^D$ and $\beta_c^B$ are attenuation coefficients related to direct signal and backscatter, respectively. Vector $v_{D}$ and  $v_{B}$ are related to the coefficients $\beta_c^D$ and $\beta_c^B$. Since multiple mapping solutions are possible from a single underwater image to clear images, underwater image enhancement is an ill-posed problem.

\begin{figure}[h]
  \centering
\vspace{+0.3cm}
\includegraphics[width=\linewidth]{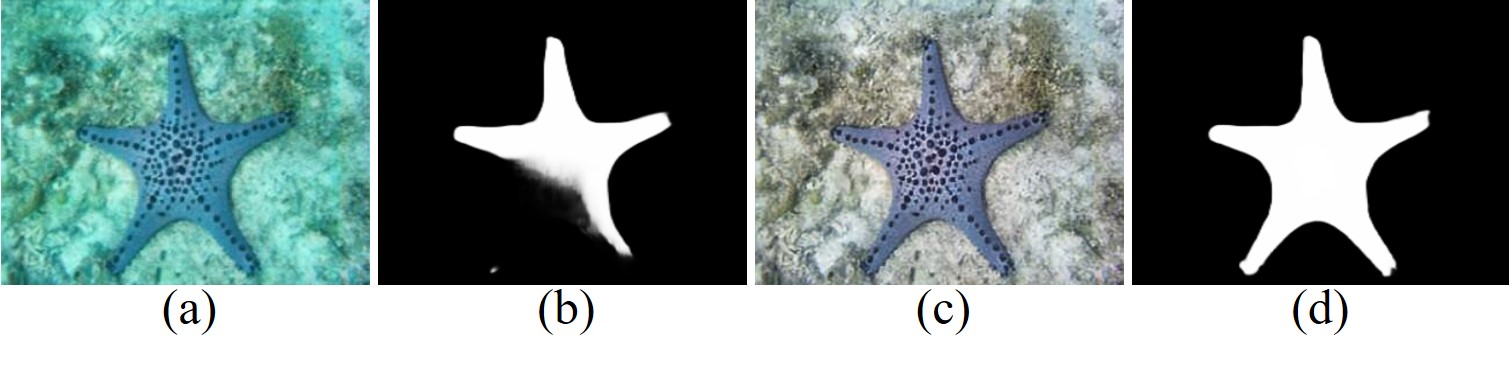}

    \caption{{Underwater image enhancement aims to reconstruct a clear image from the underwater image.
    This technique is important for autonomous underwater vehicles to implement some high-level computer vision tasks precisely. (a) Original image. (b) Salience object detection \cite{qin2019basnet} of (a). (c) Enhanced image generated by our algorithm. (d) Salience object detection of (c).
}}

\label{fig:figure2}
%\vspace{-0.5cm}
\end{figure}

Albeit of its ill-posedness, many efforts on developing visual priors capture deterministic and statistical properties of underwater images \cite{he2011single, peng2017underwater, drews2013transmission,galdran2015automatic}. These methods estimate $B_c^{\infty}$ based on specific priors and solve $J_c$ eventually.
However, utilizing these methods may result in disagreeable artifacts because their handcrafted visual priors from human assumptions cannot always hold in various real-world images. Instead of adopting handcrafted visual priors, recently, deep convolutional neural networks have been successfully used in many computer vision problems \cite{huang2021deepopht, liu2018synthesizing,yang2021multi,yang2021S3Net}, and so does underwater image \cite{fabbri2018enhancing, li2019underwater, islam2020sesr}, which achieves real improvement against conventional prior-based methods. 
Therefore, in this paper, we also develop a CNN model to tackle underwater image enhancement. The previous deep-learning-based works have obtained outstanding performance; however, they are impracticable for real-world robot applications due to limited computing resources like memory size and parameters on robotic systems.

We observe most neural network models for low-level vision like image denoising and image dehazing \cite{yang2019wavelet,chen2019pms,yang2020net} tend to employ the encoder-decoder structure that contains down-sampling and up-sampling operations. Down-sampling in the model diminishes feature maps but increases the receptive field to extract multi-scale features. Then up-sampling is applied to magnify the diminished feature maps and reconstruct clear images. Though down-sampling does not take more parameters but discard certain information. On the other hand, up-sampling methods like transposed convolution \cite{gao2019pixel} and pixel shuffle \cite{shi2016real} for precise estimation take not only more computational efforts but also more parameters in the model. To address this issue, we abandon the down-sampling and up-sampling in our model and propose an adaptive feature fusion (AFF) module. A feature map is passed through the AAF module consisting of multiple convolution kernels to obtain feature maps with multi-scale semantic information. Furthermore, the channel attention mechanism is employed to distribute the weights of three feature maps adaptively. 
Besides the AAF module, we also use residual modules \cite{he2016deep} that alleviate the difficulty of vanishing gradient in deep neural networks. Overall, our lightweight adaptive feature fusion network (LAFFNet) is based on the U-Net \cite{ronneberger2015u,yang2019causal,yang2020characterizing} and consists of these two modules. Furthermore, to decrease parameters in the model, we do not apply compact convolutions but simply reduce the channels of all convolution, and the model still achieves good performance.

We make the following contributions in this paper:
\begin{enumerate}
\item we propose the novel lightweight model contains AFF modules, which generate different scale feature maps and effectively fuse them with channel attention. The LAFFNet is a deep end-to-end trainable neural network without assuming any restrictions on attenuation coefficients and wideband veiling light. Furthermore, Our method reduces the number of parameters from 2.5M to 0.15M (around 94\% reduction). 
\item We conduct several experiments and experimental results prove that our LAFFNet achieves much more accurate performance than previous state-of-the-art methods on EUVP \cite{islam2020fast} and UFO-120 \cite{islam2020sesr} datasets.
\item  Additionally, we perform salience object detection \cite{qin2019basnet} and single image depth estimation \cite{yin2019enforcing} on our enhanced images and obtain better results, which is significant for marine robots.
\end{enumerate}

We organized our paper as follows. Section II provides brief reviews of related works like underwater image enhancement and feature fusion in neural networks. In Section III, we describe the proposed network with the AFF module and loss function Additionally, the analysis of our meteork is provided . In Section IV, experimental results of ablation studies and performance compared with conventional methods are described. We also demonstrate our model can be efficiently incorporated into other high-level vision systems to obtain better performance. Section V is the conclusion.

\section{RELATED WORKS}

\subsection{Underwater Image Enhancement}

\begin{figure*}[h]
\vspace{0.2cm}
  \centering
\includegraphics[scale=0.55]{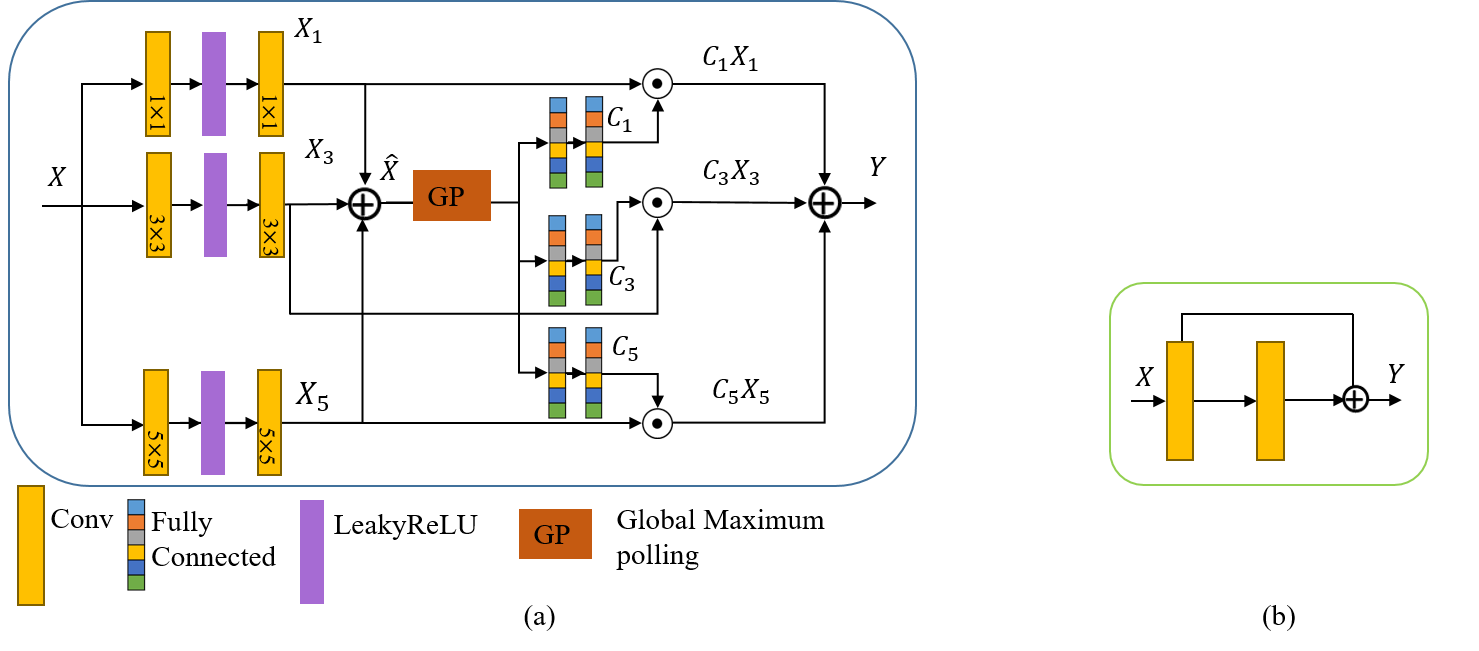}

    \caption{{ The proposed AFF module and the residual module in our model. (a) AFF module. Digits in convolution are the kernel size of each convolution. (b) the residual module.
}}

\label{fig:figure2}
%\vspace{-0.5cm}
\end{figure*}

\begin{figure*}[h]
  \centering
\includegraphics[scale=0.6]{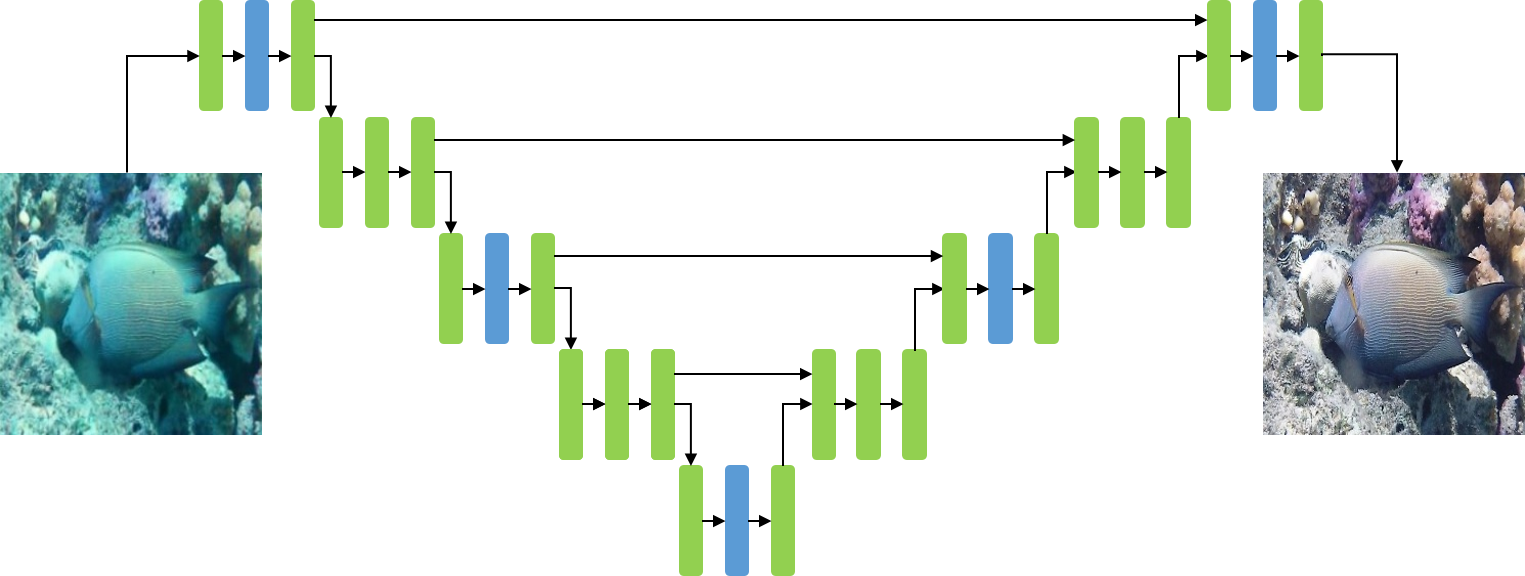}

    \caption{{ The overall LAFFNet that is based on U-Net subsumes nine local blocks. Blue blocks are AFF modules, and green blocks are residual modules. Five local blocks contain the AFF module, and the rest local blocks contain three residual modules.
}}

\label{fig:figure2}
%\vspace{-0.5cm}
\end{figure*}

\begin{table*}[]
  \centering
    \caption{{The parameters of all components in the LAFFNet.}}
\begin{tabular}{|l|l|l|l|}
\hline
Name & Input Size & Output Size & Parameters \\ \hline
$Conv^2_{1\times1}$   & $256\times256\times16$ & $256\times256\times16$ & $1\times1\times16\times16$ + $1\times1\times16\times16$\\ \hline
$Conv^2_{3\times3}$   & $256\times256\times16$ & $256\times256\times16$ & $3\times3\times16\times16$ + $3\times3\times16\times16$\\ \hline
$Conv^2_{5\times5}$   & $256\times256\times16$ & $256\times256\times16$ &  $5\times5\times16\times16$ + $5\times5\times16\times16$\\ \hline
Summation & $256\times256\times16$ & $256\times256\times16$ &0\\ \hline
GP & $256\times256\times16$ & $1\times1\times16$ &0\\ \hline
$fc_i^2$ & $1\times1\times16$ & $1\times1\times16$ &  $16\times16 + 16\times16$  \\ \hline
\end{tabular}
\vspace{-0.5cm}
\end{table*}

The methods of underwater image enhancement are divided into prior-based and learning-based. Because of the high similarity between the underwater image enhancement model and the haze model \cite{mccartney1976optics}, many methods based on dehazed models are proposed to tackle underwater images. For example, Dark Channel Prior (DCP) \cite{he2011single} is the most widely popular algorithm for image dehazing. The DCP depends on the assumption that hazy images consist of pixels that have very low intensities (close to zero) at least one color channel. In \cite{peng2015single}, blurriness prior (BP) is proposed according to the observation which the deeper the scene depth is, the more blurred the underwater object. Then BP is utilized to estimate scene depth and reconstruct clear images. Furthermore, image blurring and light absorption (IBLA) \cite{peng2017underwater} extended BP is developed to estimate more accurate underwater scene depth and background light and enhanced underwater images under different types of complicated scenes. Despite acquiring a series of success, these visual prior methods are not robust to deal with various situations like the unconstrained environment in the wild. In view of the prevailing success of deep learning in computer vision \cite{huang2021deepopht}, and robotic navigation \cite{yang2021causal} tasks and the availability of large image datasets, many deep-learning-based methods are proposed.

In \cite{li2019underwater}, WaterNet utilizes an encoder-decoder network and a novel fusion-based strategy to reconstruct a clear image from an underwater image directly. In \cite{islam2020sesr}, DEEP SESR incorporates dense residual-in-residual sub-networks to facilitate multi-scale hierarchical feature learning for both enhancement and salience object prediction. Recently, several generative adversarial networks (GAN) \cite{goodfellow2014generative} based underwater image enhancement models which generate realistic images have obtained impressive results from both unpaired \cite{fabbri2018enhancing} and paired \cite{islam2020fast} training. In \cite{fabbri2018enhancing}, Fabbri et al. propose U-GAN by the popular CycleGAN \cite{zhu2017unpaired} approach, which desires to translate an image from one arbitrary domain $X$ to another arbitrary domain $Y$ without pre-defined image pairs. In \cite{islam2020fast}, FUnIE-GAN is proposed. Authors introduce a fully-convolutional conditional GAN-based model for underwater image enhancement and formulate a multi-modal objective function to train the model. However, GAN-based methods are prone to training instability and time-consuming; hence it is necessary to tune a careful hyper-parameter. Furthermore, the generated images tend to consist of spatially inconsistent stylizations with undesirable artifacts. 
Due to these challenges, the end-to-end neural network is selected in our task. We also add various modules to improve performance.

\subsection{Feature Fusion in Neural Networks}
Feature fusion eases the difficulty of training networks with hundreds of layers and improves the robustness and accuracy of the network. 
In \cite{szegedy2015going}, Inception is proposed to summarize feature maps from different scale convolutions. 
ResNet \cite{he2016deep} introduces an identity skip connection which alleviates the difficulty of vanishing gradient in deep neural network and allows network learning deeper feature representations. DenseNet \cite{huang2017densely} strengthens feature propagation and encourages feature reuse to substantially reduce the number of parameters. 
Though these models obtain impressive improvement, these methods directly add multiple feature maps without considering the weights of each feature map. 
Thus, the attention mechanism is proposed to attend to some important parts. In \cite{hu2018squeeze}, SENet introduces a channel-attention mechanism by adaptively recalibrating the channel feature responses. 
Furthermore, the attention mechanism is widely applied for various image processing tasks like image denoising \cite{chen2018color,tsai2018efficient}, image deraining \cite{yang2020wavelet} and so on. Similarly, our AFF module contains channel attention to control the weights of different multi-scale feature maps adaptively. 

\section{Proposed methods}
Our LAFFNet is the encoder-decoder structure and subsumes two main blocks: the AFF module and the residual module. The architecture of LAFFNet is shown in Fig. 3.
We describe these two modules before elaborate on the whole network. Furthermore, we also provide a detailed analysis of our LAFFNet. Finally, the loss function is described to train the proposed model.
 
\subsection{Adaptive Feature Fusion and Residual Module} 
An AFF module is a computational unit that is constructed upon a transformation mapping an intermediate feature map $ X\in R^{H\times W\times C} $ to a feature map $ Y\in R^{H\times W\times C} $ and plotted in Fig. 1(a). Given an intermediate feature map $ X\in R^{H\times W\times C} $, the two-layered 1$\times$1, 3$\times$3 and 5$\times$5 convolution units are connected to obtain three various scale feature maps, and the relationship is written as:

\begin{equation}
%\vspace{+0.5cm}
X_{1} = Conv^2_{1\times1}(X), X_{3} = Conv^2_{3\times3}(X), X_{5} = Conv^2_{5\times5}(X)
\label{eq:res}
%\vspace{-0.5cm}
\end{equation} 
where subscripts $i \in {1, 3, 5}$ mean the size of convolutional kernels is $i\times i$. It is noted that we denote the two-layered convolution as $Conv^2$.
Three feature maps are then merged from multiple branches via an element-wise summation:
\begin{equation}
%\vspace{+0.5cm}
\hat{X} = X_{1}+X_{3}+X_{5}
\label{eq:res}
%\vspace{-0.5cm}
\end{equation} 
Then we transform $\hat X$ into the channel-wise tensor by simply using global maximum pooling and two sequential fully-connected layers for three feature maps, and the formula can be expressed as:
\begin{equation}
%\vspace{+0.5cm}
C_i = fc^2_i(GP(\hat{X}))
\label{eq:res}
%\vspace{-0.5cm}
\end{equation} 
where $GP$ is global maximum pooling, $fc^2_i$ means two sequential fully-connected layers for different feature maps, and $C_{i} \in R^{1\times 1\times C}$. The $C_1$, $C_3$, and $C_5$ are three channel-wise tensors for the precise and adaptive selections.
Finally, the output $Y$ is the summation of $C_1\otimes X_{1}$, $C_3\otimes X_{3}$ and $C_5\otimes X_{5}$ :
\begin{equation}
%\vspace{+0.5cm}
Y = C_1 \otimes X_{1}+ C_3 \otimes X_{3}+ C_5 \otimes X_{5}
\label{eq:res}
%\vspace{-0.5cm}
\end{equation} 
where $\otimes$ is channel-wise multiplication.
The details of the AAF module are listed in Table 1. This module subsumes three two-layered convolutions with different sizes, global maximum pooling, and three two-layered fully connected layers. The channel numbers of two-layered convolutions are 16, which is very lightweight.

The second component is the residual module. This module consists of two convolutional ($Conv$) layers. Given an intermediate feature map $ X\in R^{H\times W\times C} $, the final output $ Y\in R^{H\times W\times C}$ is written as:
\begin{equation}
%\vspace{+0.5cm}
Y = Conv(Conv(X))+X
\label{eq:res}
%\vspace{-0.5cm}
\end{equation} 
Comparing to the conventional CNN, residual layers are intelligently learned residual functions with reference to layer inputs, instead of learning whole functions \cite{he2016deep}. This reformulation makes the training process effective, especially in the event of deeper networks. Both modules are employed to construct the LAFFNet and shown in Fig. 2. 
\subsection{Network Architecture}
Our network is based on the U-Net that is widely used for image processing \cite{yang2019wavelet}, speech enhancement \cite{yang2020characterizing}, and so on. As shown in Fig. 3, a sequence of residual modules and AFF modules consecutively connects, which aims at learning the feature map between $I_c$ and $J_c$. Similar to U-Net, our model equips the skip-connection between corresponding structure channels, and the entire network is divided into nine local blocks. The first five local blocks are encoder parts that serve as the multi-level information extractor, and the rest four blocks are decoder parts that fuse information to reconstruct clear images. All local blocks in the LAFFNet subsume three modules. The $1^{st}$, $3^{rd}$, $5^{th}$, $7^{th}$ and $9^{th}$ local blocks consist of an AFF module between two residual modules. The rest local blocks are stacks of three residual modules. 
\noindent \smallskip\\
{\bf Analysis of the Lightweight Model:} To design the lightweight model, many methods like depth-wise convolution \cite{howard2017mobilenets} and knowledge distillation \cite{hinton2015distilling} are proposed. Nevertheless, in this paper, we do not employ these methods but just reduce the channel in convolutions. According to \cite{han2020ghostnet}, the authors indicate redundancy in feature maps is an important characteristic of those successful CNNs but increases both model parameters and computational cost. Furthermore, unlike high-level computer vision tasks, such as recognition, and detection, the underwater image enhancement task does not require the high dimensional feature. Thus, reducing the channel in the convolutions is feasible to design the lightweight network for underwater image enhancement. We empirically set the channel of both modules as 16. Our LAFFNet takes 0.15M parameter, and GFLOPs is 9.77, which is lightweight and compact for underwater robotic application.

\subsection{Loss Function}
To train our LAFFNet, we apply three loss functions.
First, existing approaches have proven that adding an $L_1$ loss to the
loss function enables the network to learn to sample from a globally similar space in an $L_1$ sense \cite{isola2017image}. In our implementation, the Charbonnier loss \cite{barron2019general}, that is a robust $L_1$ loss, is selected as the objective function and expressed as:
\begin{equation}
%\vspace{+0.5cm}
L_{Cha}(J_{c},\hat{J}_{c})= \sqrt[]{(J_{c}-\hat{J}_{c})^2+\epsilon ^2}
\label{eq:res}
%\vspace{-0.5cm}
\end{equation}
where $J_c$ and $\hat J_c$ mean the ground truth and predicted clear images, respectively, and $\epsilon$ is a very tiny constant (e.g., $10^{-3}$).
This loss function is robust to handle outliers and stable during training. It is noted when $\epsilon$ is 0, Eq.(7) is an $L1$ loss.
Second, local structures and details are important factors to be taken into consideration while enhancing underwater images. To measure these factors, the similarity index (SSIM) loss proposed by \cite{zhao2016loss} is added to the objective function. SSIM of $x$ and $y$ is defined as:

\begin{equation}
  {\rm SSIM}(x,y) = \frac{(2\mu_x\mu_y + C_1)  (2 \sigma _{xy} + C_2)} 
    {(\mu_x^2 + \mu_y^2+C_1) (\sigma_x^2 + \sigma_y^2+C_2) }
  \label{eq:SSMI}
\end{equation}
where {$\mu$} and {$\sigma$} represent the means, standard deviation, and covariance of images, while $C_1$ and $C_2$ are the variables to stabilize the division. The loss function for the SSIM can be written as follows:
\begin{equation}
  L_{{SSIM}}(J_{c},\hat{J}_{c}) = 1-{\rm SSIM}(J_{c},\hat{J}_{c})
  \label{eq:SSMIloss}
\end{equation}
The third loss is perceptual loss \cite{johnson2016perceptual} that reconstructs images with high-level semantic features. The perceptual loss is expressed as:

\begin{equation}
%\vspace{+0.5cm}
L_{Per}(J_{c},\hat{J}_{c})= |(VGG(J_{c})-VGG(\hat{J}_{c})|
\label{eq:res}
%\vspace{-0.5cm}
\end{equation}
where $VGG$ is the classical VGG-19 \cite{simonyan2014very} network and $| \cdot |$ is the absolute value. Combining $L_{Cha}$, $L_{SSIM}$ and, $L_{Per}$ the overall loss function is written as:
\begin{equation}
%\vspace{+0.5cm}
L_{Total}= \lambda_{1} L_{cha}+ \lambda_{2} L_{SSIM}+ \lambda_{3} L_{Per}
\label{eq:res}
%\vspace{-0.5cm}
\end{equation}
where $\lambda_{1}$, $\lambda_{2}$ and $\lambda_{3}$ are scaling coefficients to adjust the importance of the respective loss components. In practice, we tune their values as hyper-parameters.

\begin{figure*}[h]
\vspace{0.4cm}
  \centering
\includegraphics[width=\linewidth]{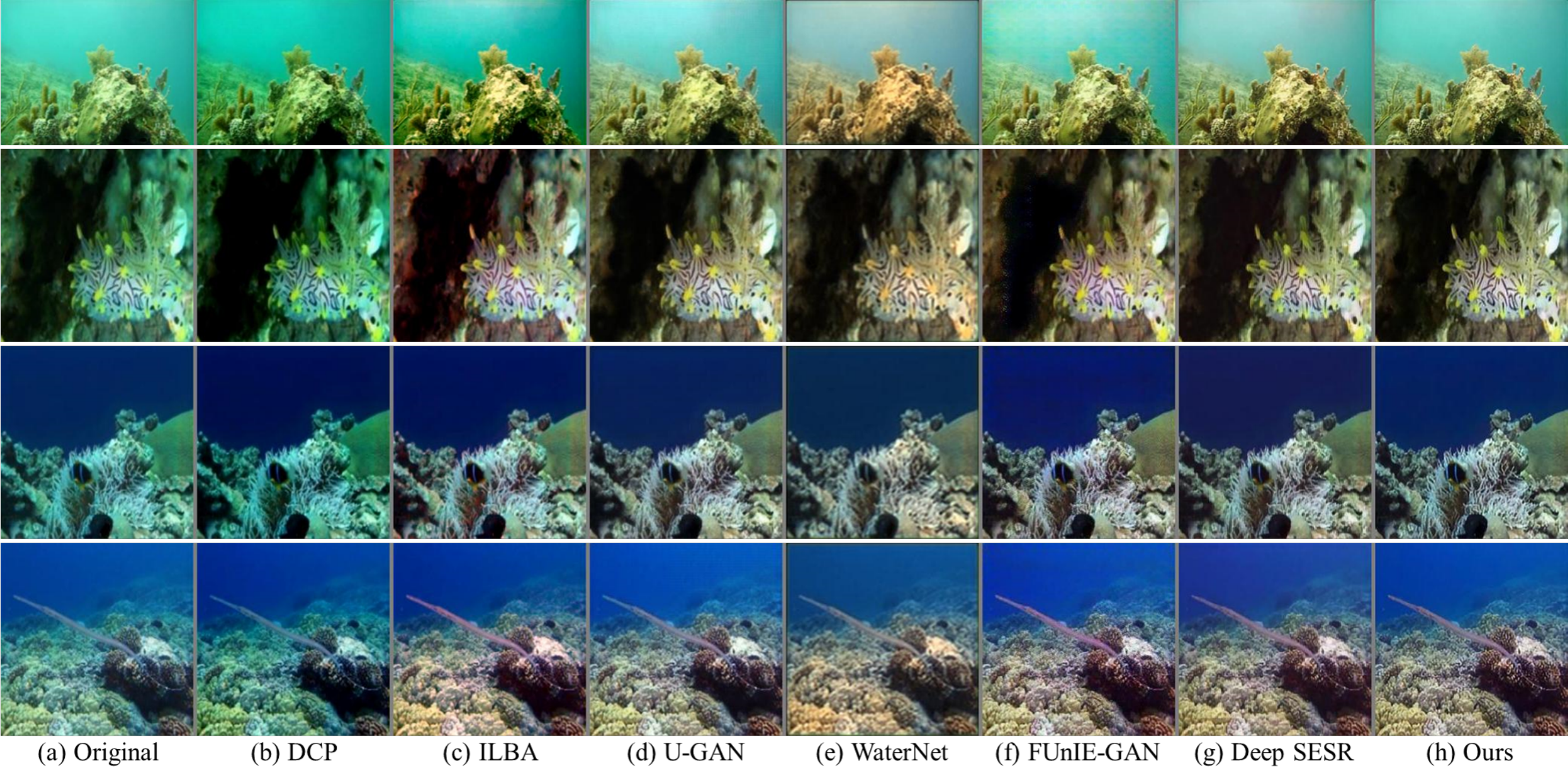}
\vspace{-0.8cm}
    \caption{{ Qualitative comparison for underwater image enhancement performance of LAFFNet with existing solutions and SOTA models on the EUVP and UFO-120 datasets: DCP \cite{he2011single}, ILBA \cite{peng2017underwater}, U-GAN \cite{fabbri2018enhancing}, WaterNet \cite{li2019underwater}, FUnIE-GAN \cite{islam2020fast} and Deep SESR \cite{islam2020sesr}.
}}

\label{fig:figure2}
%\vspace{-0.5cm}
\end{figure*}
\section{Experimental results}
\subsection{Dataset and Experimental Setup}

In this work, we adopt the UFO-120 \cite{islam2020sesr} and EUVP \cite{islam2020fast} datasets for evaluation. 
The UFO-120 dataset contains over 1500 samples for training and validation, and another 120 for testing. On the other hand, there are 12K training and validation samples and 515 samples for evaluation in the EUVP dataset. During training, images are resized to 240$\times$320 and 256$\times$256, which follows the protocol in \cite{islam2020sesr}.$\lambda_{1}$, $\lambda_{2}$ and $\lambda_{3}$ are set 1, 1.1 and 0.1 in our experiments. Adam \cite{kingma2014adam} is used as an optimization algorithm with a mini-batch size of 5. We set the initial learning rate as 0.001 and divide it by 10 after 30 epochs. The models are trained for 200 iterations. For the EUVP dataset, we adopt the model trained on UFO-120 and fine-tune it. We implement entire experiments by the PyTorch framework and train on NVIDIA GeForce GTX 2080 graphics cards.

\subsection{Underwater Image Enhancement Results}

Peak Signal-to-Noise Ratio (PSNR), Structural Similarity (SSIM), and Underwater Image Quality Measure (UIQM) \cite{panetta2015human} are chosen as objective metrics for quantitative evaluation. We select six state-of-the-art works to make fair comparisons with our method. The six comparative methods are DCP \cite{he2011single}, IBLA \cite{peng2017underwater}, Water-Net \cite{li2019underwater}, U-GAN \cite{fabbri2018enhancing}, FUnIE-GAN \cite{islam2020fast} and Deep SESR \cite{islam2020sesr}. The first two methods are prior-based methods, and the others are deep-learning-based methods, as introduced in Section II. The model parameter, GFLOPS, average PSNR and SSIM are presented in Table II. For convenience, some metrics are cited from \cite{islam2020sesr}. The proposed method outperforms the state-of-the-art methods. Furthermore, our model takes the least parameters and GFLOPs, which demonstrates feature fusion is beneficial for lightweight networks.

\begin{table*}[h]
\vspace{0.2cm}
    \centering
    \caption{{The model parameter, GFLOPS, average PSNR and SSIM values of enhanced results on the EUVP and UFO-120 datasets. We represent the best two results in {\color{red} red} and {\color{blue} blue} colors. PSNR and SSIM scores are shown as $mean\pm \sqrt {variance}$.}}
    \begin{tabular}{|c|c|c|c|c|c|c|}
    \hline
    \multirow{2}[4]{*}{} & \multirow{2}[4]{*}{GFLOPs} & \multirow{2}[4]{*}{\# Model param} & \multicolumn{2}{c|}{EUVP} & \multicolumn{2}{c|}{UFO120} \\
\cline{4-7}          &       &       & PSNR  & SSIM  & PSNR  & SSIM \\
    \hline
DCP \cite{he2011single} &   &   & $17.55\pm2.8$ &$0.69\pm0.07$ & $18.20\pm3.1$& $0.71\pm0.06$   \\ \hline
IBLA \cite{peng2017underwater}    &  & &$18.83\pm4.5$ &$0.70\pm0.15$ & $17.50\pm5.2$& $0.65\pm0.17$  \\ \hline
WaterNet \cite{li2019underwater}   & 142.9 & {\color{blue}1.09M} & $20.14\pm2.3$ &    $0.68\pm0.18$  &$22.46\pm1.9$ & $0.79\pm0.05$   \\ \hline
U-GAN  \cite{fabbri2018enhancing} & 18.14 & 38.7M    &   $23.67\pm1.5$    &   $0.67\pm0.11$  &$23.45\pm3.1$  & $0.80\pm0.08$  \\ \hline
FUnIE-GAN \cite{islam2020fast}  & \color{blue}10.23  &7.01M    &   ${ \color{blue}26.78\pm1.1}$&    ${ \color{blue}0.86\pm0.05}$  &$25.15\pm2.3$&  $0.82\pm0.08$  \\ \hline
Deep SESR \cite{islam2020sesr} & 146.1  &2.46M&$25.25\pm2.1$ &    $0.75\pm0.07$ & ${ \color{blue}27.15\pm 3.2}$ &${ \color{blue}0.84\pm0.03}$\\ \hline
Ours & {\color{red}9.771} & {\color{red}0.15M} &${ \color{red} 28.42\pm4.0}$ &${ \color{red} 0.87\pm0.07}$ & ${ \color{red} 28.94\pm2.6}$& ${ \color{red} 0.86\pm0.04}$  \\ \hline
\end{tabular}

\end{table*}

Some enhanced images are plotted in Fig. 4. Enhancement results estimated by prior based methods \cite{he2011single} \cite{peng2017underwater} fall short in hue rectification and contain some color distortions in some regions. 
Compared with state-of-the-art, the proposed method has the best performance in terms of water removal and artifact/distortion suppression (see, e.g., Fig. 4 (h)).

Furthermore, we calculate the average UIQM of enhanced images. UIQM is the linear combination of UICM, UISM, and UIConM \cite{panetta2015human}. UICM, which quantifies the degradation caused by light absorption, is defined by the statistics of the differences between red-green and yellow-blue planes. UISM depends on the strength of Sobel edges computed on each colour channel independently; whereas UIConM is calculated by the logAMEE operation \cite{panetta2010parameterized}, which is considered consistent with human visual perception in low light conditions. The relationship of four metrics is written as:
\vspace{-0.3cm}
\begin{equation}
%\vspace{-0.3cm}
\rm{UIQM}= 0.028\times \rm{UICM}+0.295\times \rm{UISM}+ 3.375 \times \rm{UIConM}
\label{eq:res}
%\vspace{-0.5cm}
\end{equation}
 We list all average metrics of enhanced images in Table III. Though the proposed network does not obtain the best performance on UICM and UIConM, our model has the highest UIQM overall.
\begin{table}[]
  \centering
    \caption{{average uicm, uism, uiconm and uiqm results on enhanced images.}}
\begin{tabular}{|l|c|c|c|c|}
\hline
   & UICM & UISM & UIConM & UIQM \\ \hline
DCP \cite{he2011single} &   \color{blue}6.781    &   4.005   &   0.056&  1.575    \\ \hline
ILBA \cite{peng2017underwater}& 7.892&  4.389    &  0.123 & 1.958\\ \hline
U-GAN \cite{fabbri2018enhancing} &  6.052    &  5.120    &   0.224&  2.483    \\ \hline
WaterNet \cite{li2019underwater} &  \color{red}6.736    &  5.292    &   0.212&  2.511    \\ \hline
FUnIE-GAN \cite{islam2020fast} & 7.040    & \color{blue}5.606    &   0.185&  2.514    \\ \hline
Deep SESR \cite{islam2020sesr} &  5.975    & 5.211&   \color{red}0.260&  \color{blue}2.638    \\ \hline
Ours &  6.502    &   \color{red}6.724   & \color{blue}0.258 &  \color{red}3.092  \\ \hline
\end{tabular}
\end{table}

\subsection{Ablation Study}
We analyze the advantage of the residual module, the AFF module, and the perceptual loss. The experimental results tested on UFO-120 are shown in Table IV. Firstly, in index-1, we replace proposed modules with vanilla convolutions. Its performance is unsatisfying and only achieves 25.89 PSNR score and 0.84 SSIM score. Secondly, in index-2, when using residual modules, the performance is improved by 2.41 and up to 28.30. Thirdly, in index-3, the extra AFF modules are added, and the PSNR score is improved by 0.47. Finally, in index-4, we add perceptual loss to our objective function; the PSNR and SSIM are further improved by 0.17 and 0.01, respectively. This demonstrates the effectiveness of our modules and perceptual loss.
\begin{table}[h]
  \centering
    \caption{{The ablation study shows the effeteness of the residual module, the AFF module, and the perceptual loss.}}
\begin{tabular}{|c|c|c|c|c|c|}
\hline
Index&Res & AFF & Per&PSNR & SSIM \\ \hline
1   &&&&25.89  &   0.84\\ \hline
2  & $\surd$  &&&28.30  &   0.85\\ \hline
3& $\surd$   &  $\surd$  & &   28.77  &   0.85  \\ \hline
4&  $\surd$   &  $\surd$    &$\surd$ &  28.94  &  0.86   \\ \hline
\end{tabular}
\end{table}
\subsection{Pre-processing for High-level Vision Tasks}
\begin{figure}[t]
  \centering
  \vspace{+0.8cm}
\includegraphics[width=\linewidth]{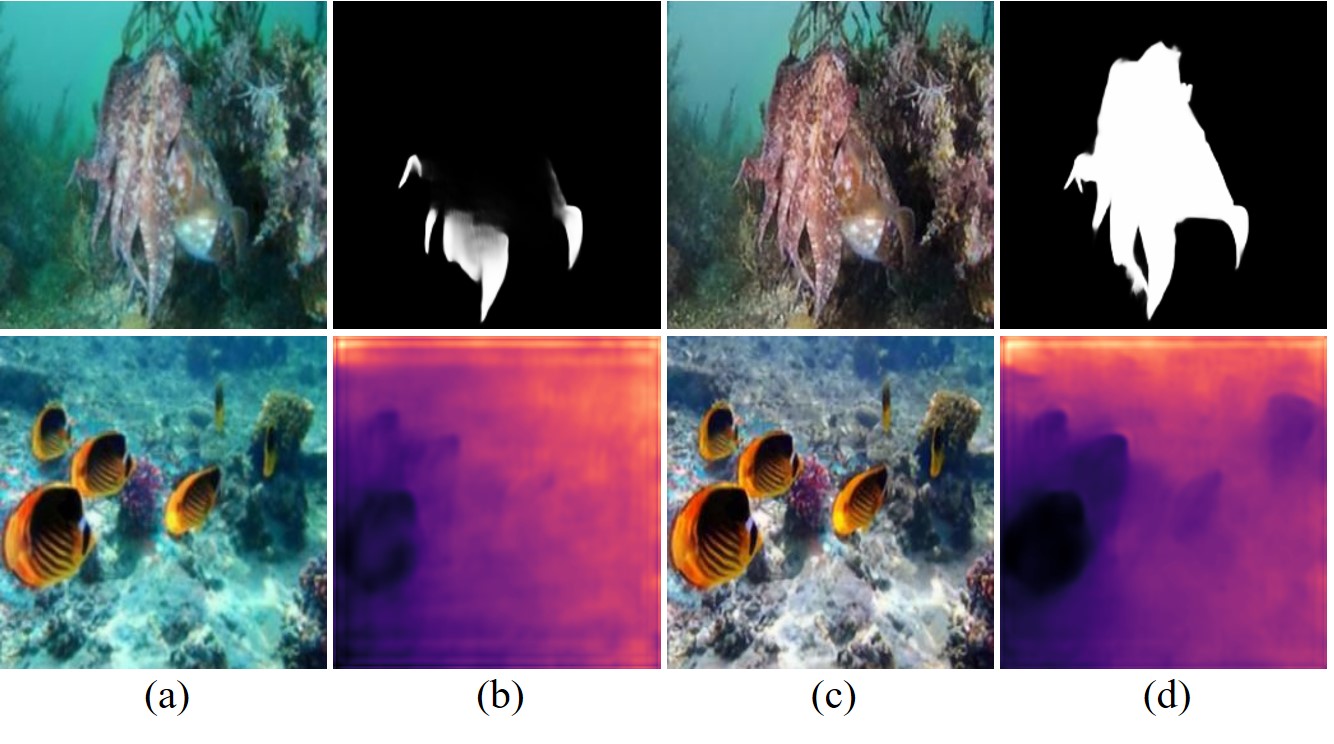}
%\vspace{-0.8cm}
    \caption{{Examples of underwater image enhancement for salience object detection and single image depth estimation on real-world underwater images. (a) Original images. (b) Salience object detection and single depth image estimation result on (a). (c) Enhanced images by our LAFFNet. (d) Salience object detection and single depth image estimation result on (c).
}}

\label{fig:figure2}
\vspace{-0.5cm}
\end{figure}
Due to the lightweight architecture, our LAFFNet can potentially be incorporated into other high-level vision systems. For example, we study problems of salience object detection \cite{qin2019basnet} and single image depth estimation \cite{yin2019enforcing} in underwater environments. 
Because underwater images can blur objects and scenes, the performance of salience object detection and single image depth estimation degrades in the water. Fig. 5 shows the visual results of salience object detection by combining with the BASNet \cite{qin2019basnet}, and single image depth estimation by \cite{yin2019enforcing}. It is obvious that the green and blue hue degrades the performance of the two tasks. For example, the predicted depth on original images cannot separate the fish and the background. On the other hand, the performance of salience detection and depth estimation on enhanced images provide a significant improvement over identical models.

\section{CONCLUSIONS}
This work introduces a lightweight underwater image enhancement method for constrained computing resources in marine robots called LAFFNet. This model abandons down-sampling and up-sampling and contains adaptive feature fusion modules to extract multi-scale features and aggregate them by channel attention. We also decrease the channel of convolutions to design the lightweight model, so the overall model has approximately 0.15M parameters, which is less and faster than other state-of-the-art models. Several experimental results on multiple datasets \cite{islam2020fast,islam2020sesr} present that our LAFFNet outperforms other state-of-the-art methods. We also implement ablation studies to show the contributions of each proposed module. Moreover, due to the generality and lightweight architecture, we demonstrate our LAFFNet to improve high-level vision tasks like salience object detection and single image depth estimation. In the future, we will combine our model with other autonomous underwater vehicles and remotely operated vehicle to implement complex tasks. Furthermore, we will investigate the ability of our model on different enhanced tasks like image desnowing \cite{chen2020jstasr} and image dehazing \cite{chen2020pmhld} for different robotic applications.

\bibliographystyle{ieeetr}
\bibliography{ref}

\end{document}